\documentclass[lettersize,journal]{IEEEtran}
\usepackage{amsmath,amsfonts}
\usepackage{algorithmic}
\usepackage{array}
\usepackage[caption=false,font=normalsize,labelfont=sf,textfont=sf]{subfig}
\usepackage{textcomp}
\usepackage{stfloats}
\usepackage{url}
\usepackage{verbatim}
\usepackage{graphicx}
\usepackage{lipsum}
\usepackage{authblk}
\usepackage{cite}
\usepackage{float}
\usepackage{balance}
\usepackage{marvosym}
\begin{document}
%\title{OmniTester:Generate and test on one click based on LLM. }
\title{Multimodal Large Language Model Driven Scenario Testing for Autonomous Vehicles}
%\author{Qiujing Lu, Xuanhan Wang, Yiwei Jiang, Guangming Zhao, Mingyue Ma, Shuo Feng\textsuperscript{\Letter} 
%Department of Automation, Tsinghua University\\
%{\tt\small \{brx22, li-s23, hehl21\}@mails.tsinghua.edu.cn} \\
%{\tt\small \{qiujinglu, fshuo\}@mail.tsinghua.edu.cn}} 

\author[1]{Qiujing Lu}
\author[1]{Xuanhan Wang}
\author[1]{Yiwei Jiang}
\author[2]{Guangming Zhao}
\author[2]{Mingyue Ma}
\author[1]{Shuo Feng}
\affil[1]{Department of Automation, Tsinghua University}
\affil[2]{Research Institute for Road Safety of the Ministry of Public Security, Beijing, China} 

%\markboth{Journal of \LaTeX\ Class Files,~Vol.~18, No.~9, September~2020}%
%{How to Use the IEEEtran \LaTeX \ Templates}

\maketitle

\begin{abstract}
The generation of corner cases has become increasingly crucial for efficiently testing autonomous vehicles prior to road deployment. However, existing methods struggle to accommodate diverse testing requirements and often lack the ability to generalize to unseen situations, thereby reducing the convenience and usability of the generated scenarios. A method that facilitates easily controllable scenario generation for efficient autonomous vehicles (AV) testing with realistic and challenging situations is greatly needed. To address this, we proposed OmniTester: a multimodal Large Language Model (LLM) based framework that fully leverages the extensive world knowledge and reasoning capabilities of LLMs. OmniTester is designed to generate realistic and diverse scenarios within a simulation environment, offering a robust solution for testing and evaluating AVs. In addition to prompt engineering, we employ tools from Simulation of Urban Mobility to simplify the complexity of codes generated by LLMs. Furthermore, we incorporate Retrieval-Augmented Generation and a self-improvement mechanism to enhance the LLM's understanding of scenarios, thereby increasing its ability to produce more realistic scenes. In the experiments, we demonstrated the controllability and realism of our approaches in generating three types of challenging and complex scenarios. Additionally, we showcased its effectiveness in reconstructing new scenarios described in crash report, driven by the generalization capability of LLMs. 
 %It can generate general cases or modified maps based on the provided map library.
\end{abstract}


\section{Introduction}
\label{sec:intro}
%\IEEEPARstart{W}{elcome} 
%\textbf{Some background about scene generation}
\IEEEPARstart{S}cenario-based testing is vital in the development of autonomous driving, as it allows AVs to be tested in specially crafted scenarios inside simulation. This method is essential for evaluating performance, identifying weaknesses, and ensuring safety— key factors in assessing AVs before road testing. However, the scenario libraries currently in use, primarily sourced from real-world data, fall short due to the scarcity of corner cases. To make things worse, as the intelligence level of autonomous driving improves, the rarity of critical events becomes increasingly problematic, exacerbating what is referred to as the "curse of rarity" \cite{liu2024curse}. The insufficient testing coverage and inefficiencies throughout the testing process is hindering enhancements in AV safety. As such, the efficient generation of testing scenarios is of critical importance. Various techniques has been explored, such as dense deep reinforcement learning \cite{feng2023dense} and adversarial framework \cite{wang2021advsim}. Yet, considering the diverse testing requirements and different development stages in AV development, effectively designing and generating suitable scenarios to meet the testing needs remains an unresolved challenge.

Many efforts have been made in this area, primarily focusing on generating challenging scenarios based on predefined functions or well-defined search algorithms. However, there has been limited exploration into developing an effective control mechanism that allows flexible management of scenario generation based on requirement descriptions. This is particularly important because developers often conceptualize scenarios in abstract terms while simulations require precise configurations for execution. For instance, developers might envision turning scenarios, while simulation requires detailed road geometry, precise initial vehicle placements and behaviors for each turn. Enhancing the controllability of the scenario generation system based on abstract descriptions can bridge the gap between developers and scenario-based testing, making it a more viable tool and expediting the path to efficient performance evaluation for AV systems. 

However, building such text-conditioned generation mechanism is challenging as it demands modeling capacity ranging from static elements in road structures to agent behaviors and map between narrative language to detailed configurations. Additionally, it must ensure the conformity to user requests, maintain he diversity of the road network, and preserve the fidelity of the generated scenarios. Such generation process requires a high level of intelligence, as it involves not only understanding the control signal but also reasoning from the request to design then generate the desired testing scenarios. The emergence of LLMs and Vision-Language Models (VLM), trained with vast amounts of data from the internet, has demonstrated remarkable intelligence, encompassing learning, reasoning, linguistic capabilities, human-like communication, and complex thinking abilities. There have been extensive explorations into their applications in fields such as medicine, education, finance, and engineering. Notably, OpenAI's CodeX \cite{chen2021evaluating}  and DeepMind's AlphaCode \cite{li2022competition} has demonstrated promising coding capabilities, MathPrompter \cite{imani2023mathprompter} has shown advancements in mathematical reasoning, and SceneCraft \cite{kumaran2023scenecraft} has excelled in creating narrative experiences.

Motivated by these impressive advancements in large language models (LLMs), this paper aims to explore how the coding, narrative, and reasoning capacities of LLMs, along with the extensive knowledge acquired from the internet, can be harnessed to bridge the gap between developers and the simulation systems. Building such a simulation tool with a robust text-conditioned mechanism presents several challenges: Firstly, the realistic and accurate generation of road networks and vehicle movements from text is inherently complex and demanding. Unlike generating vivid images or videos from text, as seen with SORA \cite{videoworldsimulators2024} and Kling \cite{kling}, this task demands a higher degree of accuracy beyond mere visual realism. The challenge lies in generating domain-specific, complex topological structures that adhere to precise spatial relationships. Secondly, hallucinations and errors inherent in LLMs \cite{xiao2021hallucination} can lead to failures in generating reasonable road networks or meaningful vehicle dynamics, especially in less common testing scenarios. Ensuring precise control based on text input remains a significant challenge and requires further research. Thirdly, given that LLMs are trained on general corpora, a significant challenge is how to make LLMs quickly understand generation requests for new scenarios and produce specific outputs while continuously learning with minimal supplied information.

\begin{figure}
  \centering
\includegraphics[width=0.4\textwidth]{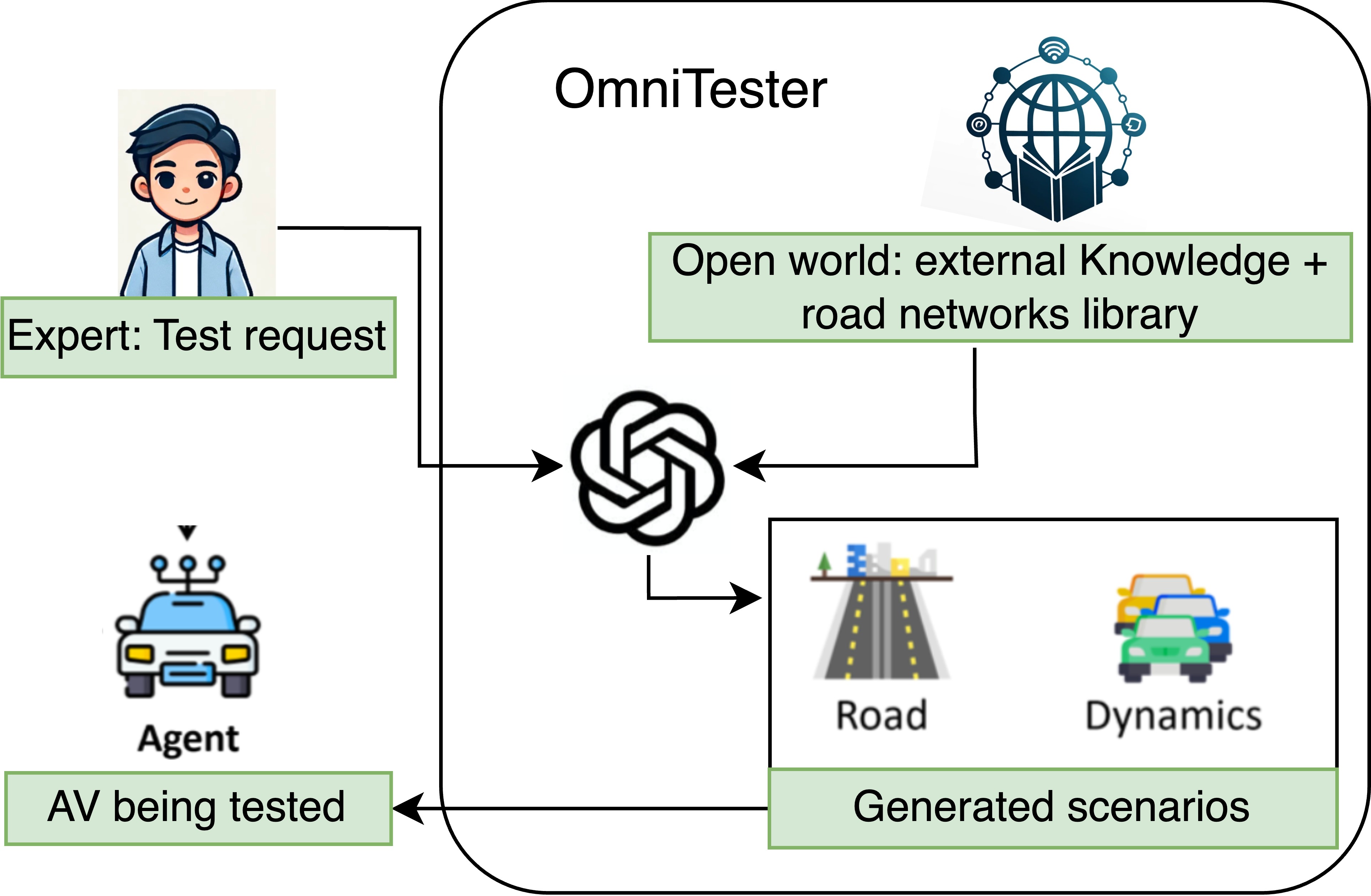}
  \caption{The LLM generation framework of OmniTester}
\label{fig:intro}
\end{figure}

To address these challenges and fully harness the inherent intelligent capabilities of LLMs, we introduced OmniTester as shown in Fig.\ref{fig:intro}, which proposes a fully automated pipeline that responds to user requests, effectively generating the desired scenarios to test the targeted functionality of autonomous vehicles. OmniTester generates realistic and diverse scenarios containing road geometries that closely mimic real-world environments through prompt engineering and the integration of tools from Simulation of Urban Mobility(SUMO), an open-source traffic simulation package \cite{lopezMicroscopicTrafficSimulation2018}. Meanwhile, OmniTester uses self-improvement mechanisms designed to continuously enhance LLM performance, reducing hallucination and inherent errors. It also uses a Retrieval-Augmented Generation (RAG)-based generation mechanism \cite{lewisRetrievalAugmentedGenerationKnowledgeIntensive2020} that automatically queries and extracts external knowledge to adapt to the operational design domain requirements. To the best of our knowledge, OmniTester is the first system to generate road structures and vehicles solely based on user requests, offering exceptional controllability and flexibility, powered by RAG and self-improvement techniques.

The rest of this paper is organized as follows: related works on scenario generation and the application of LLMs in the autonomous driving domain is reviewed in Section \ref{sec:related_work}. Section \ref{sec:method} details the design and implementation of OmniTester, with a focus on the prompt engineering techniques and RAG implementation. In the section \ref{sec:result}, we showcase the various scenarios generated by OmniTester, and evaluate the effectiveness of the system's key components. Additionally, we include a case study featuring scenario generation from a crash report as an application of our system. Finally, conclusions are presented in Section \ref{sec:conclusion}.

\begin{figure*}[ht!]
  \centering
\includegraphics[width=0.95\textwidth]{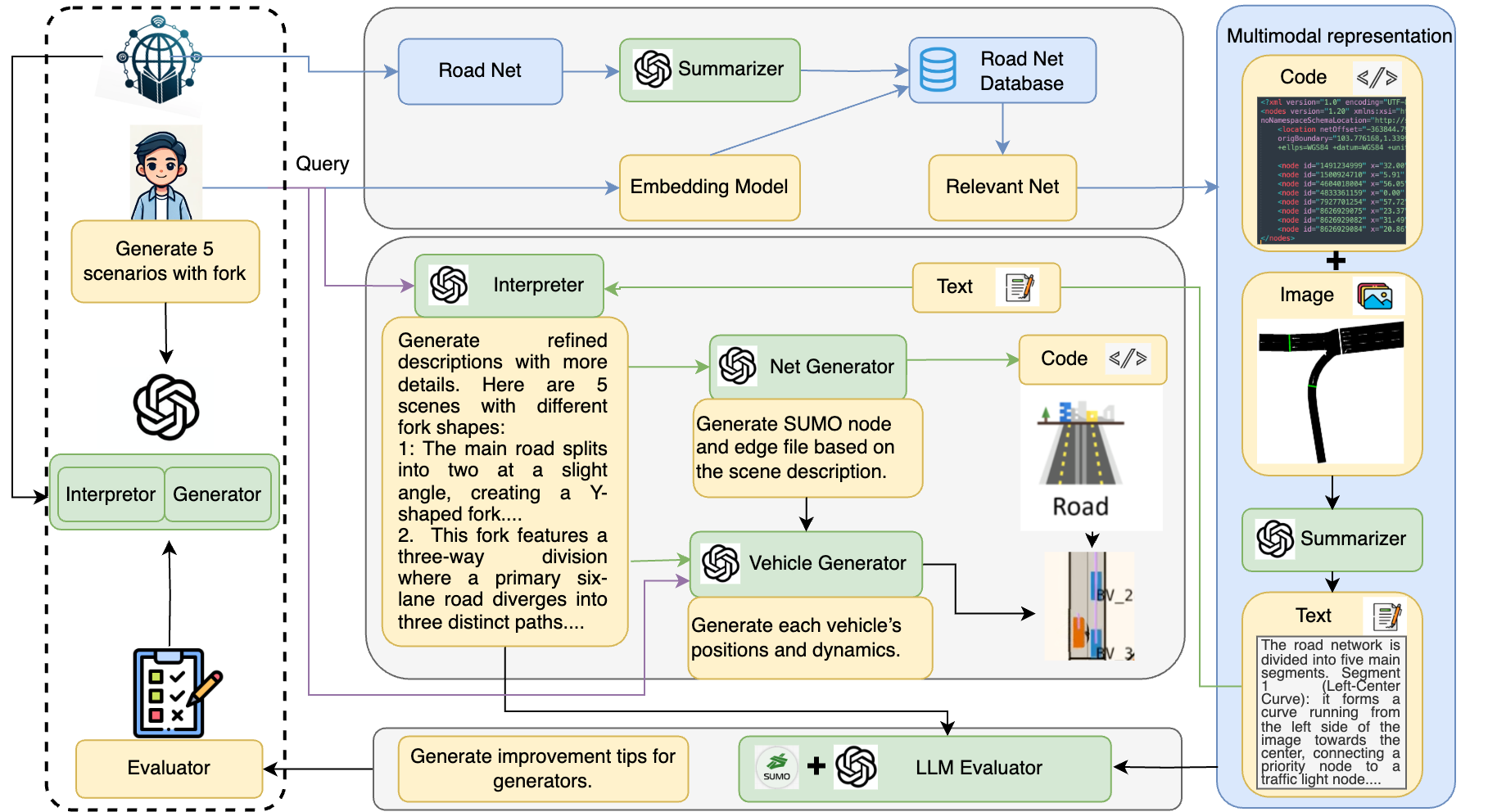}
  \caption{Dataflow within OmniTester: The Interpreter, RAG module, Net Generator, Vehicle Generator and LLM Evaluator are activated upon the user's request.}
\label{fig:framework}
\end{figure*}

\section{RELATED WORKS}\label{sec:related_work}
\subsection{Scenario-based testing}
Safety is the major factor holding back the widespread deployment of AVs, extensive efforts have been made to identify and address unsafe components through rigorous testing \cite{zhangPerformanceEvaluationMethod2022,zhaoGeneticAlgorithmBasedSOTIF2023,fahrenkrogEuropeanResearchProject2023}. Compared to road testing, scenario-based testing aims to offer more efficient, targeted assessments with better coverage of corner cases \cite{nalic2020scenario}. This approach tests AVs in specially crafted scenarios that highlight safety-critical situations. However, designing and generating these scenarios remains challenging, and various methods have been developed to address this. One straightforward method is replaying the logged behavior of all vehicles in the scene, which ensures realistic behavior for each agent. Nevertheless, this method has limitations because the entire environment cannot adapt to the new behaviors of AVs, and it is restricted to existing collected scenarios, resulting in unreliable evaluation outcomes with limited coverage.

Various methods have been explored for generating novel challenging scenarios for testing, such as combination-based, worst-case, adaptive scenario generation. The combination-based scenario generation approach \cite{zhou2017reduced} decomposes scenarios into several basic scenario units and constructs complex scenarios by permuting and combining these basic units. The worst-case scenario generation approach \cite{jung2007worst} uses the human steering angle as a design variable, optimizing it to produce the most challenging scenarios for AVs. The adaptive scenario generation approach \cite{mullins2018adaptive} calibrates the AV surrogate model adaptively to explore and determine the model's performance boundaries, generating representative scenarios for safety evaluation. However, these methods can only provide a limited range of scenarios and fail to model the agents within these scenarios realistically. As a result, they do not capture the full complexity of real-world situations.

 Data-driven models have also been explored for scenario building, where they are used to model key components of the scenario, ranging from agent behaviors to the road network within the simulation. For instance, NeuralNDE \cite{yan2023learning} uses a Transformer-based network with safety mapping to provide realistic agent behaviors, achieving distributional-level similarity to real-world distributions. STRVE \cite{Rempe_Philion_Guibas_Fidler_Litany_2022} learns a graph-based conditional VAE as traffic prior, optimizing each agent's behavior to provoke collisions with a rule-based AV planner. SLEDGE \cite{chitta2024sledge} employs Diffusion Transformers to jointly generate lanes and agents, serving as the initial state for traffic simulation. RealGen \cite{ding2023realgen} uses an encoder-decoder architecture and retrieval-based in-context learning to synthesize realistic traffic scenarios. However, these models primarily focus on generating realistic scenarios based on provided datasets, the challenge of synthesizing novel, challenging situations and effectively controlling the generated context for specific testing purposes remains unsolved.

\subsection{Scenario Generation with LLMs}
Multimodal large language models have been utilized in multiple aspects of autonomous driving systems \cite{cui2024survey}.  \cite{fu2024drive} \cite{sha2023languagempc}  \cite{xu2023drivegpt4} \cite{wang2023drivemlm} explored the application of LLM (GPT3.5/4) and VLM (GPT4) on the autonomous driving systems. These works seek to leverage the reasoning, interpretive, and memorization capabilities of large pretrained models to comprehend the driving environment in a human-like manner and to make informed driving decisions when confronted with complex scenarios. Similarly, multimodal large language models have been utilized for scenario generation. ChatScene \cite{zhang2024chatscene} uses LLM to query a given database written in Scenic code \cite{fremont2019scenic} for generating scenarios based on user queries. By utilizing the reasoning and understanding capabilities of LLMs, it achieves fine performance. However, the Scenic code snippets that contribute to the main scene setup are drawn from predefined library, limiting its generalization ability and effectiveness in generating new and challenging scenes. ChatSim \cite{wei2024editable} utilizes LLMs to collaboratively edit and generate photorealistic 3D driving scene simulations through natural language commands, enabling efficient and interactive scene modifications. However, it limits modifications to existing scenes, lacking the capability to generate entirely new scenes from scratch.

LEADE \cite{tian2024llm} proposed a scenario generation approach that utilizes LLM-enhanced adaptive evolutionary search to create safety-critical and diverse test scenarios for autonomous driving systems testing. However, it fails to model safety-violation scenarios caused by anomalous actions from agents. CTG++ \cite{zhong2023language} introduces the use of the LLM to transform a user's query about safety-critical scenarios into the corresponding differentiable loss function of a diffusion model to generate the query-compliant trajectories. However, it can only manipulate agent behaviors within a given road map with specified initial locations. LLMScenario \cite{chang2024llmscenario} utilizes LLM to generate short trajectories for agents based on minimal scenario descriptions, facilitating scenario engineering. However, its application is currently limited to highway scenarios, and further exploration is needed to extend its use to more complex environments.

As a significant and evolving area of research, the challenge of enhancing controllability and generalization in generating complex situations remain inadequately addressed. Our work aims to address this challenge by generating road structures and vehicles from scratch with a multimodal LLM-driven, text-conditioned pipeline.

\section{Method}
\label{sec:method}
In this section, we delineate the OmniTester system, the multimodal LLM-driven tool designed for text-conditioned scenario generation. Within this system, we introduce several techniques to enhance LLM's understanding and generation capabilities for scenarios. We begin by introducing the overall generation framework, followed by an explanation of the prompt engineering techniques utilized for scenario generation. Subsequently, we provide an in-depth explanation of the Net Generator and Vehicle Generator components. Finally, we introduce the RAG mechanism.
\subsection{Pipeline}
\begin{figure*}[htbp] 
  \centering  
\includegraphics[width=0.98\textwidth]{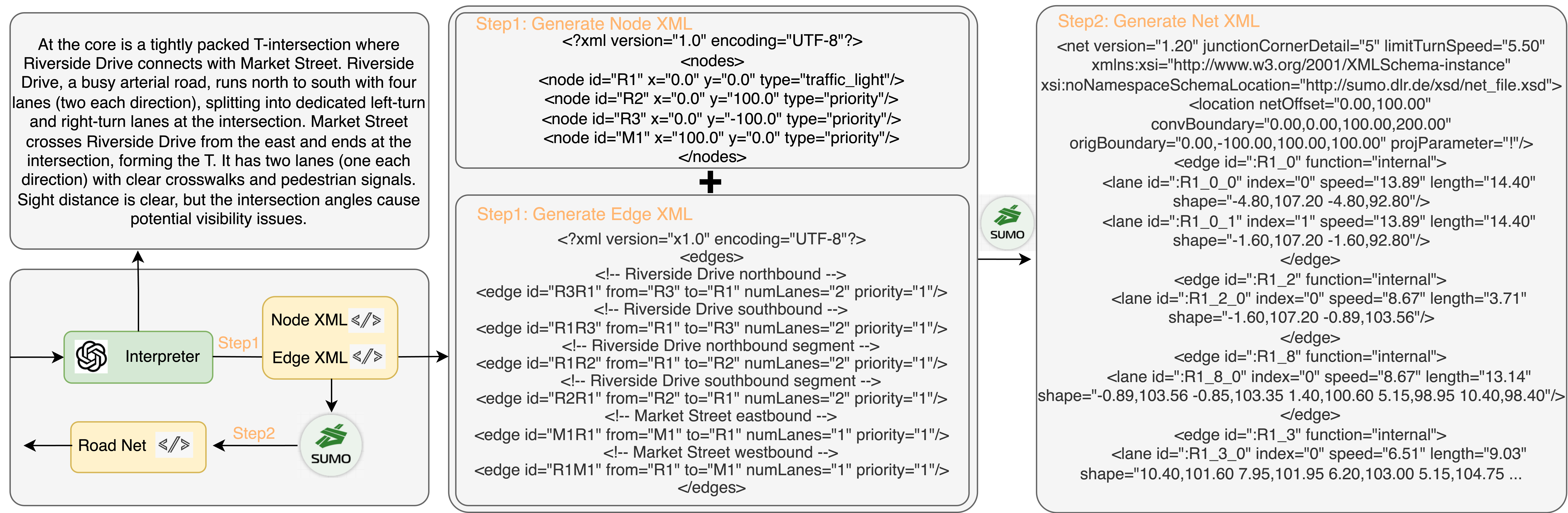}
  \caption{
Generate a road XML file through a two-step process: first, use a properly prompted LLM to directly generate the node and edge files, then use SUMO to convert these into the corresponding net file, all in XML format. \label{fig:road}}  
\end{figure*}
The generation framework is illustrated in Fig. \ref{fig:framework}. The process of creating a set of scenarios starts when the user submits specific testing requests, such as "Generate 5 scenarios with a fork." Then an LLM is invoked to elaborate on the scenario details through descriptive means, such as "the main road splits into two at a slight angle.. " for the 5 scenarios requested. Once the scenario descriptions are complete, a Net Generator, powered by the LLM, constructs a corresponding road network to match the outlined scenarios. After the road network is successfully generated, another LLM agent is used to configure the vehicle dynamics involved in the scenarios.

Subsequently, an evaluator based on LLM is deployed to assess whether the scenario generated aligns with the user's specified testing intention, focusing on background vehicle (BV) behaviors. At the same time, a RAG module can be optionally activated to refine the LLM's comprehension of the intended road category. In instances where the behavior of the AV or BV does not meet the desired criteria, feedback is directly incorporated into the prompt. This prompts the LLM to regenerate the scenario, incorporating the evaluator's reasoning in the revised output. This iterative process ensures that the generated scenarios are both accurate and consistent with the user's initial testing objectives.

\subsection{Scenario Prompt Engineering}

The LLM's ability to comprehend the scenarios both geometrically and programmatically, will greatly impact the controllability and realism of the generated scenarios. We utilized various prompt techniques to integrate domain knowledge, generate more structured outputs, and maximize the reasoning capabilities of LLMs, while minimizing the hallucination. Chain-of-Thought (CoT) prompting \cite{wei2022chain} motivates the LLM to articulate its reasoning process before providing the final answer. We designed specific prompts to break down the generation task into a series of subproblems, leading the model to tackle these sequentially to achieve the ultimate solution. For example, in network generation, we utilize a multi-step process to aid the model in understanding user requests. Initially, we direct the model to summarize the task in the “Description” section and then guide it to "step by step explain your reasoning process" in the “Reasoning” section. Refer to the prompt example in Appendix 1. This method not only helps the LLM to better comprehend and analyze the testing requirements but also encourages rational thought with minimal hallucination. For further evidence of the importance of this component, see the ablation study in section \ref{sec:ablation}. 

Secondly, we specify the output requirements for the LLM. We instruct the LLM to generate the correct format of node and edge files by providing detailed formatting guidelines. Additionally, we impose additional constraints on road length and design rules to enhance the realism of the output network. 

Considering the complexity of output, the varying number of vehicles, and the heterogeneous information (such as edge and time information), we use the few-shot prompting technique  \cite{brown2020language} to help the model understand the context and generate the desired format of the response with a higher success rate. Please refer to the Appendix \ref{sec:appendix} for detailed prompt samples.

\subsection{LLM-driven Scenario Generation}
We decompose scenario generation into two stages: road network generation and agents' routes generation. Road structure is important since testing on accurate and varied road structures ensures that AVs comply with road safety regulations and can respond appropriately to diverse road conditions. In our implementation for road generation, a properly prompted LLM-based Interpreter is employed to generate a detailed description for the whole scenario, then SUMO compatible Node and Edge file defined in XML format is generated based on the description.. Lastly, SUMO tool is utilized to convert them into single net file defined in XML format, representing the entire road network. See Fig. \ref{fig:road} for an illustration and more detailed examples in Section \ref{sec:result}. This design is not confined to SUMO or its XML formats. Since road network naturally represents a graph structure, it can be represented by other structured languages \cite{GraphMLFileFormat} and processed by graph tools \cite{hagbergExploringNetworkStructure2008}, compatible to other simulators such as MATSim \cite{MultiAgentTransportSimulation}.

Considering close interactions with other vehicles are among the most challenging scenarios for AVs \cite{IntelligentDrivingIntelligence}, we utilize LLM to create such scenarios by generating specific routes for the vehicles involved. The intelligence of LLMs is leveraged to set up suitable spatial and temporal positions of vehicles and AVs within a given scenario. Specifically, LLM is called with specially designed prompts to generate trips for the desired number of vehicles as described by the interpreter. This generator is tailored to generate AV and BV's departing and arriving edges as well as departure time, according to the test goal and the road structure. Once these parameters are established, detailed routes, including all the edges the vehicle will pass, are computed using the duarouter tool \cite{sumo_duarouter}, which optimizes the routes based on the road network and traffic conditions.

\begin{figure}
\centering 
\includegraphics[width=0.46\textwidth]{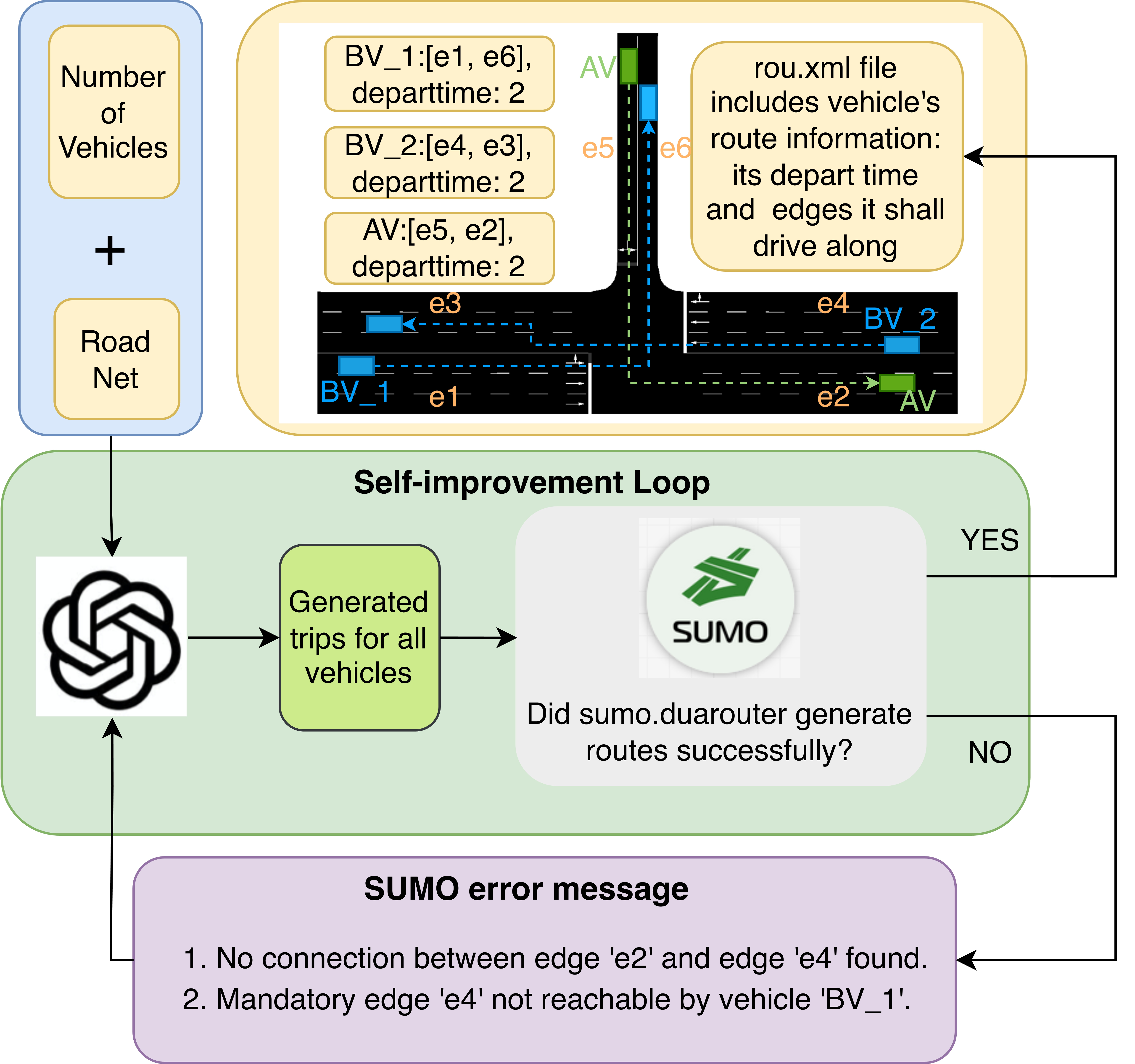}
\caption{Self-improvement feedback loops for route generation. \label{fig:self-improve}} 
\end{figure}

An additional LLM is used to efficiently correct unreachable routes initially generated by the vehicle generator. Through joint debugging with the Network Conversion tool netconvert \cite{NetconvertSUMODocumentation}, the error messages provided by netconvert offer detailed instructions on what went wrong and what can be corrected. Based on the detailed feedback, the LLM-based generator updates the trips generated for vehicles. See Fig.\ref{fig:self-improve} for the detailed pipeline applied. With the help of this self-improvement mechanism, all the routes can be configured successfully.

\subsection{RAG for scenarios}
%We use Openstreetmap to download road geometry for any interested region. Once net xml is generated, a BEV image is created from sumo-gui and sent to LLM to generate descriptions for road information.  For each generated description, ada? is used for extracting embedding for it and stored into the database.  During retrieval process, the description for target scenario is first sent to ada? for embedding and similarity is compared to retrieve similar road geometry. The node and edge file is sent as additional prompt into the scenario generator for in-context learning.  
The process begins with the extraction of road geometry from OpenStreetMap \cite{OpenStreetMapFoundation}, a collaborative project that provides freely accessible, editable geographic data globally. This platform is utilized to obtain detailed and up-to-date road geometries of specified regions of interest, serving as the foundational dataset for providing open-world knowledge about road structures. The extracted road geometries are then converted into a network model using netconvert. This model is represented in the net.xml format, which is compatible with SUMO.

Following the network model creation, a bird's-eye view (BEV) image of the traffic scenario is generated using the graphical user interface of SUMO (sumo-gui). This image captures the layout and other relevant road network attributes. The BEV image serves as an input to an LLM, which processes the visual information and generates descriptive text detailing the road geometry and network characteristics. This step is critical as it bridges the gap between visual data and textual representation, facilitating easier interpretation and further processing of network characteristics in textual form. Once each textual description generated by the LLM, it is transformed into vector embeddings using OpenAI's embedding tools, text-embedding-ada-002 \cite{neelakantan2022text}. It converts the text into a high-dimensional space that captures the semantic features of the text numerically. These embeddings are then stored in a database, based on Chroma DB \cite{chromadb}, enabling efficient retrieval and comparison based on semantic similarity.

During the retrieval phase, the description of the target scenario is converted into an embedding using the same embedding tool. This embedding serves as the basis for identifying the most semantically similar road geometry stored in the database, achieved through comparison of embeddings. This method of similarity assessment ensures that the selected road geometry closely aligns with the specific requirements of the target scenario. See Section \ref{sec:result} for more details.

Once the appropriate road geometry is identified, the corresponding node and edge files are utilized as additional inputs (prompts) in the scenario generation process. This approach leverages in-context learning, where the scenario generator adjusts and refines its output by incorporating the specific contextual details provided by the node and edge files. This integration enhances the accuracy and relevance of the generated traffic scenarios, tailored to the particular characteristics of the road network.

\section{Experiments}\label{sec:result}

\begin{figure*}[htbp] 
  \centering  
\includegraphics[width=0.98\textwidth]{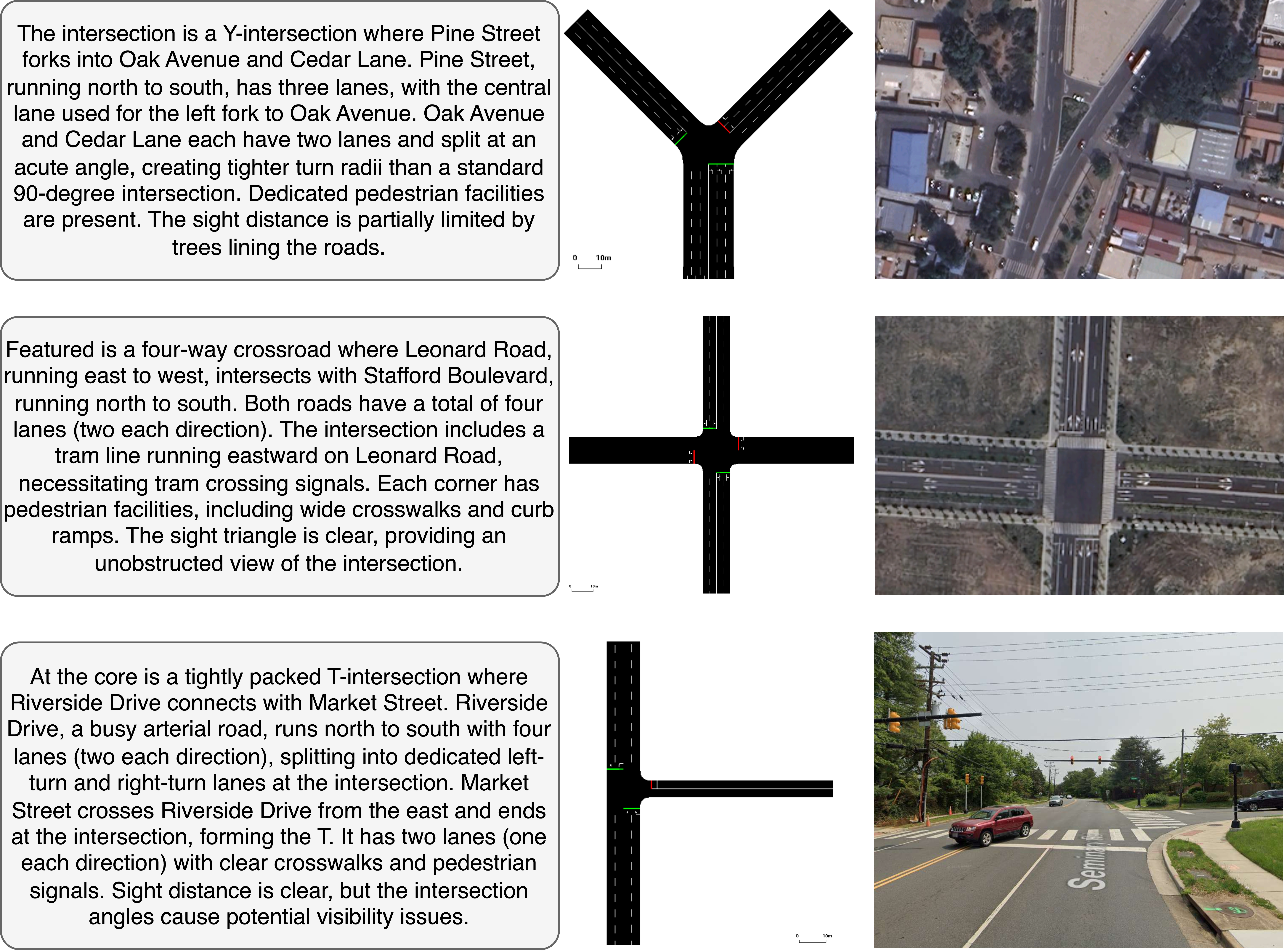}
  \caption{
  Sampled realistic intersections generated by OmniTester are presented. Left: Description generated by Interpreter; Middle: SUMO visualized road structure from the net XML file; Right: Similar roads found in the real world. \label{fig:real_road}}  
\end{figure*}

In this section, we demonstrate the effectiveness of our Multimodal LLM framework for scenario generation. Firstly, we show that the system can generate diverse scenarios to cover various situations while maintaining high realism, based on user's testing requests. Secondly, we highlight the difficulty of these scenarios by testing an LLM-driven AV within them. Thirdly, we showcase the effectiveness of RAG module. To demonstrate our system's controllability and generalization capabilities in generating scenarios with more detailed descriptions, we introduce a case study where our OmniTester generates similar risky situations purely based on text descriptions from crash reports. Lastly, we validate the importance of the key design elements in this framework through a thorough ablation study.

\subsection{Controllable realistic scenarios}\label{subsec:compliance}

Controllable scenario generation requires high conformality to user requests and realistic outputs. In our experiments, we observed that with proper prompting, the LLM demonstrates strong intelligence and high performance in this task. One detailed example of a generated road network, along with comprehensive descriptions provided by the interpreter, can be seen in Fig.\ref{fig:road}. The interpreter successfully produces detailed descriptions, including the intersection type (T-shape), road layout, and number of lanes for each direction, as well as the road name and traffic conditions for the roads. Based on this description, Net Generator intelligently places nodes to match the details, and generates edges defined on top of these nodes with lanes and connections that align with the description. As illustrated, it not only connects the nodes correctly to create a layout that matches the description but also generates comments describing the edges' direction and corresponding names for human readability.

In evaluating the performance of OmniTester systematically, we first measure the controllability of the whole system, which can be divided into two logical levels, road structure and agents related descriptions. We evaluated how the generated results are matching the user request and interpreter's descriptions from these two perspectives. Several accuracy metrics are computed for a quantitative evaluation of the alignment between requests and generated results.

The accuracy of the scene type evaluates whether the generated network's scene category aligns with the user's request or the detailed scene description provided by the interpreter. The number of lanes/vehicles measures whether the specified count in the description matches the generated result. 

We strictly count one generation pass as failure if the output does not follow required output format or generated net cannot be recognized by SUMO. Success rate is averaged over number of scenarios. Three different testing requests, each with 10 generate scenarios, are compared: general road scenarios, intersections, and fork scenarios. As seen from Table \ref{tab:conformance}, OmniTester has a high success rate in generating accurate types of road networks. Additionally, it achieves 100\% accuracy in generating the desired number of vehicles according to the interpreter's description for most cases. However, some confusion regarding the exact number of lanes for a specified road shown up when generating road networks, which might be due to inherent hallucinations in the language model \cite{xu2024hallucination}. Moreover, the generation failures are mainly due to incorrect formats in the keywords (like extra "\#" tags or additional ":"), leading to the inability to parse the node and edge files from the long text response. Considering the complexity of generating these files, the success rate for a single-pass generation is reasonable high, and correct outputs often appear through regenerations.

Next, we calculate the mean and standard deviation of the number of total lanes, edges, route length as well as number of vehicles in the generated scenario to measure the complexity of generated scenarios. As seen from Table \ref{tab:diversity}, these metrics span a wide range across different types of roads, indicating diverse scenario generation. Furthermore, compared to fork scenarios, intersections exhibit greater variations in route length and the number of lanes, which is reasonable given the complexity and diverse types of intersections. A detailed scatter plot for edges and total length can also be seen in the left figure of Fig.\ref{fig:net_scatter}, illustrating that they span a wide range and cover diverse situations.

Some sampled results are visualized in Fig.\ref{fig:real_road}. For the request of generating intersection scenarios, LLM produces typical Y-shape (top), four-way (Middle) and T-shape (Bottom) intersections. As shown, Interpreter not only generates the layout clearly, from the general direction to main segments, but also provides very detailed information such as pedestrian facility, curb ramps and sight distances so on.  In the case of the Y-shape intersection, the generated description includes critical features, specifically noting the "split at an acute angle" and providing turn radii information. This level of detail ensures that the physical layout and functional aspects are accurately represented. For the four-way intersection, the interpreter describes the moving directions of the intersecting roads with precision, ensuring that the layout is practical and aligns with typical four-way intersection configurations. This includes information on lane assignments and possible traffic flows, which are essential for realistic scenario generation. Regarding the T-shape intersection, the interpreter covers traffic conditions and splitting directions comprehensively. It describes how the traffic flows from the main road into the side road and vice versa, including details about signal controls. The final net xml outputs from SUMO in these cases match the description from the layout to the number of lanes in each segment. As shown, OmniTester is capable of generating detailed narratives for intersections with high realism and can also produce road structures on demand.

\begin{table}[!h]
\caption{Conformity of command\label{tab:conformance}}
\centering
\begin{tabular}{|c|c|c|c|}
\hline
Accuracy & General Scenarios & Intersection &  Fork\\
\hline
Scene Type & 1  & 0.9 & 1 \\
\hline
Number of lanes &1  & 1 & 0.9 \\
\hline
Number of vehicles &1   & 1 & 0.9 \\
\hline
Success rate & 0.7  & 0.8 & 0.8 \\
\hline
\end{tabular}
\end{table}

\begin{table}[!h]
\caption{Diversity\label{tab:diversity}of generated scenarios}
\centering
\begin{tabular}{|c|c|c|c|}
\hline
Scenario & General Scenarios & Intersection &  Fork\\
\hline
\#Lanes & $19.3 \pm 8.93$  & $23.50\pm 8.91$ & $10.50 \pm 5.25$ \\
\hline
\#Edges & $10.80 \pm 2.75$  & $12.0\pm 4.71$ & $6.60 \pm 2.80$ \\
\hline
Route Length& $335.98 \pm 132.22$  & $386.98\pm 162.2$ & $297.03 \pm 142$ \\
\hline
\#Vehicles & $10.6 \pm 5.43$ & $6.9\pm 2.95$ & $7.3 \pm 4.33$ \\
\hline
\end{tabular}
\end{table}

\subsection{Controllable challenging scenarios}

RandomTrip \cite{sumo2018} from the SUMO tool generates a set of random trips for a given network by randomly selecting  source and destination edges for vehicles appearing at a specified arrival rate. We used it as a baseline since it provides sufficient coverage of possible scenarios. To ensure a fair comparison with the LLM-generated scenarios, we use a slightly higher arrival rate and then randomly delete vehicles to match the number of vehicles in the generated scenarios, maintaining a consistent level of crowding. 

To evaluate the challenging level of the generated scenarios, an LLM-based AV is used to navigate inside these scenarios. The poorer its performance, the more challenging the scenarios are considered. We adopt the performance metrics from Limism++ \cite{fu2024limsim++} to quantifying the AV's performance. The driving score is computed as weighted score for ride comfort, driving efficiency, and driving safety. The route completion value is the ratio of the completed route length by the driver agent to the total route length of the preset route. For more details, see Limism++.

\begin{table}[!h]
\caption{Challenges posed by LLM based vehicle generator for fork scenario \label{tab:llm-av}}
\centering
\begin{tabular}{|c|c|c|}
\hline
Scenario & Ours & RandomTrip  \\
\hline
Route completion & $0.42 \pm 0.50$  & $0.72 \pm 0.46$  \\
\hline
Driving score &$40.22 \pm 36.54$ & $63.34 \pm 27.43$ \\
\hline
Total score &$31.09 \pm 38.74$  & $55.14 \pm 37.26$  \\
\hline
Use Time & $67.90 \pm 36.33$  & $89.71 \pm 76.56$\\
\hline
Success rate & $0.50 \pm 0.53$  &  $0.80 \pm 0.42$  \\
\hline
\end{tabular}
\end{table}

As shown in Table  \ref{tab:llm-av}, our LLM-based vehicle generator creates more challenging BV routes, resulting in significantly lower success rates and performance scores for AVs. 
\begin{figure}[H]
\centering 
\includegraphics[width=0.45\textwidth]{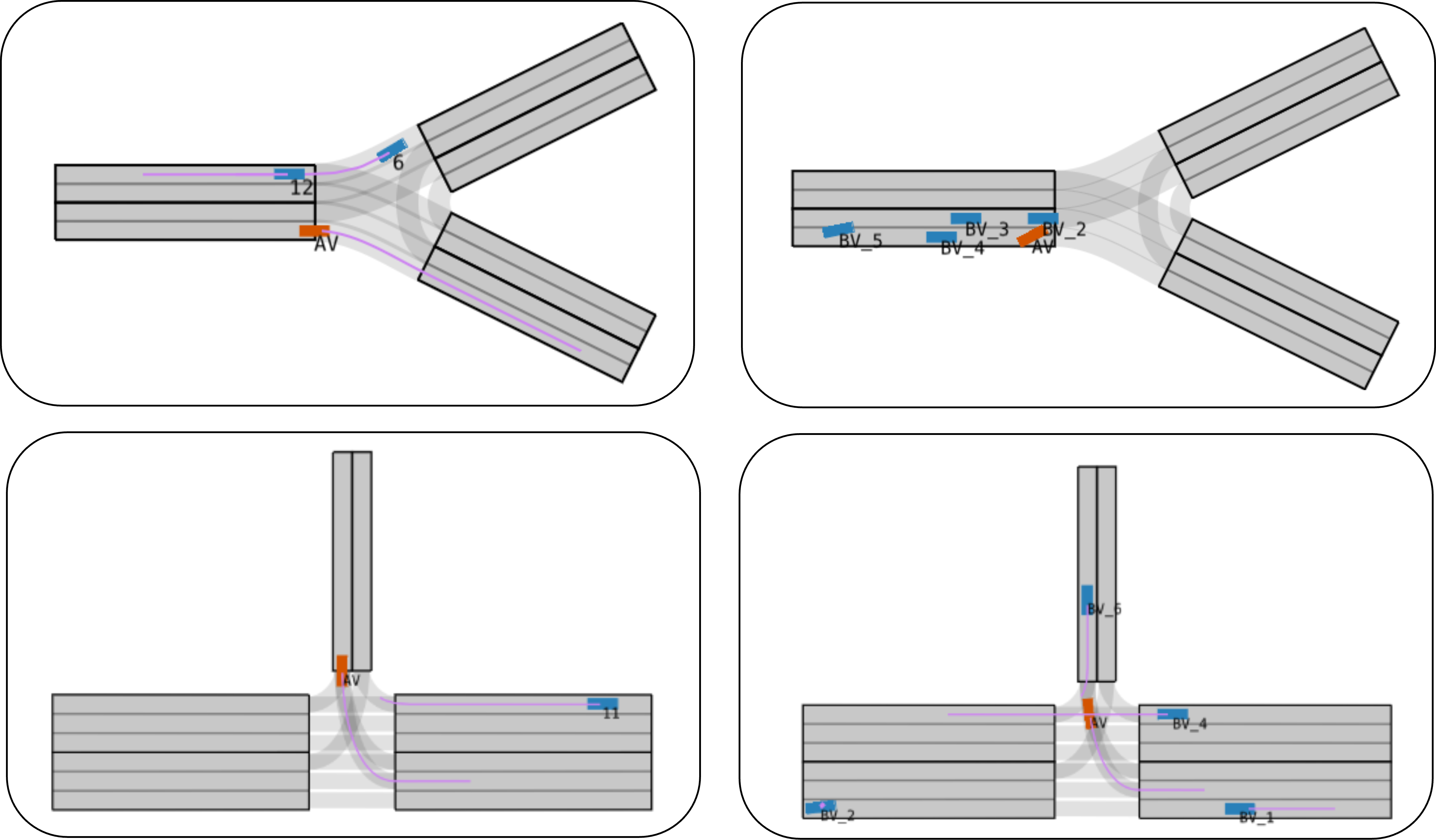}
\caption{Left: Scenarios generated by RandomTrip, where AV and BV appear randomly from entrances at random times. Right: Challenging situation created by OmniTester, where AV and BV routes are meticulously designed to generate dense interactions around AV and near the junction.
\label{fig:av-demo}} 
\end{figure}
Two sampled test cases for forking and merging situations are shown in Fig.\ref{fig:av-demo}. In the top scenario, within the same junction, the randomly distributed vehicles leave an empty lane for AV to drive through, with no other vehicles in nearby lanes. AV is passing the junction smoothly with no interaction with other vehicles, making this testing scenario less effective in testing interaction capabilities. In contrast, OmniTester creates a crowded traffic situation near the junction, requiring AV to perform a lane change before passing through. This challenging scenario ultimately results in a collision with BV2, revealing inherent flaws in the lane change strategies of this LLM-driven AV during dense traffic conditions. In the bottom scenario, at the T-shaped junction, the left-turning AV encounters no incoming vehicles to interact with in the RandomTrip setup. In contrast, OmniTester creates a complex interaction situation with a car following AV, a vehicle leading AV, and another vehicle driving towards the intersection as AV is turning. This complexity challenges AV's reasoning and decision-making processes, making it easier to reveal inherent flaws in this LLM-driven AV.

\subsection{Effectiveness of RAG Module}
\begin{figure*}[htbp]
\centering 
\includegraphics[width=0.98\textwidth]{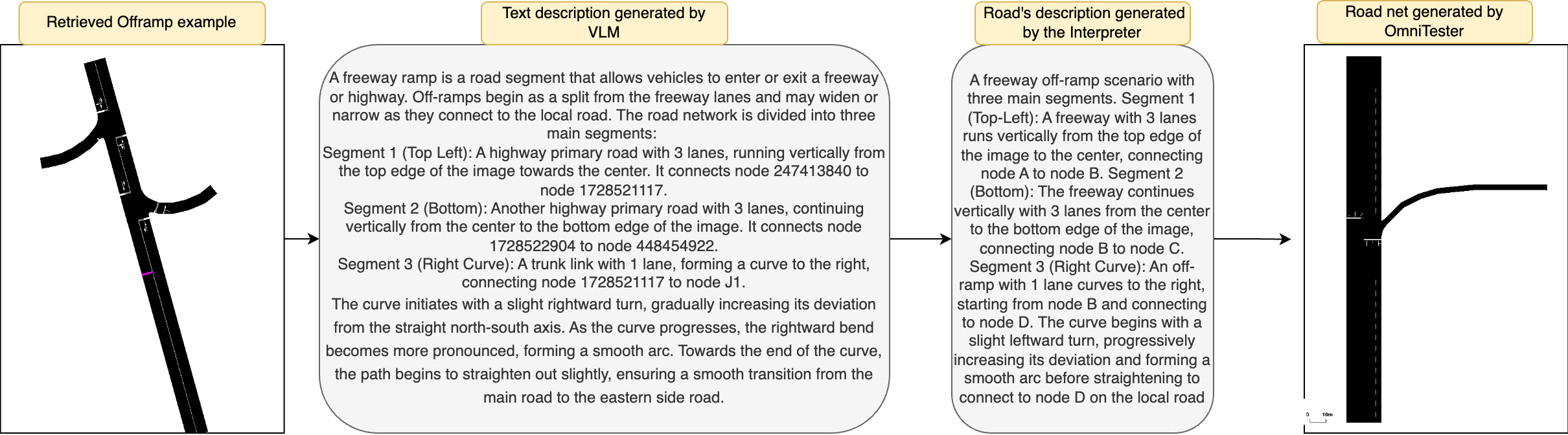}
\caption{Left: Freeway Off-ramp retrieved from database
Right: Generated road network with a similar layout. \label{fig:rag-demo}} 
\end{figure*}
To generate diverse freeway ramp scenario, we pulled several on/off ramp map from OpenStreetMap and added into the database. Upon user request, it can  generate realistic ramp scenarios with assistance from the data retrieved from this database. Instead of using code, text, or picture formats of the road network on the fly for assisting generation, a detailed description of the road structure is prepared to provide distilled information. The generated description ensures that the Net Generator receives sufficient information to construct a similar new road structure that matches the layout, segment details, and directional flow of the original. 

As illustrated in Fig.\ref{fig:framework}, the text description is generated using a VLM-based summarizer. Powered by VLM's vision and spatial reasoning capabilities, layout information can be extracted from the picture. The node and edge files offer detailed information about the road structure, including length, number of lanes, and the shape of each curved edge. With this abundant information, the summarizer intelligently selects key features based on knowledge from the open world and hints injected from the prompt. It then summarizes all the important information into the final description, which contains the general layout, main segment locations, moving directions, connectivity, and even detailed descriptions of curve shapes. Fig.\ref{fig:rag-demo} contains an example illustrating how the retrieved example's description guided the generation of freeway off-ramp road structures. As shown, it describes a freeway off-ramp comprising three segments, covering the main freeway road and the trunk for exiting the freeway. Besides detailing the layout, segments, and number of lanes for each edge, it provides a thorough description of the curve shape. Using this example in the prompt, the Interpreter can generate a similar description for the Net Generator, preserving the detailed curve information. The final generated network shares a similar shape to the provided example, especially the curvature of the offramp.

\begin{figure*}[htbp]  
  \centering  
\includegraphics[width=0.99\textwidth]{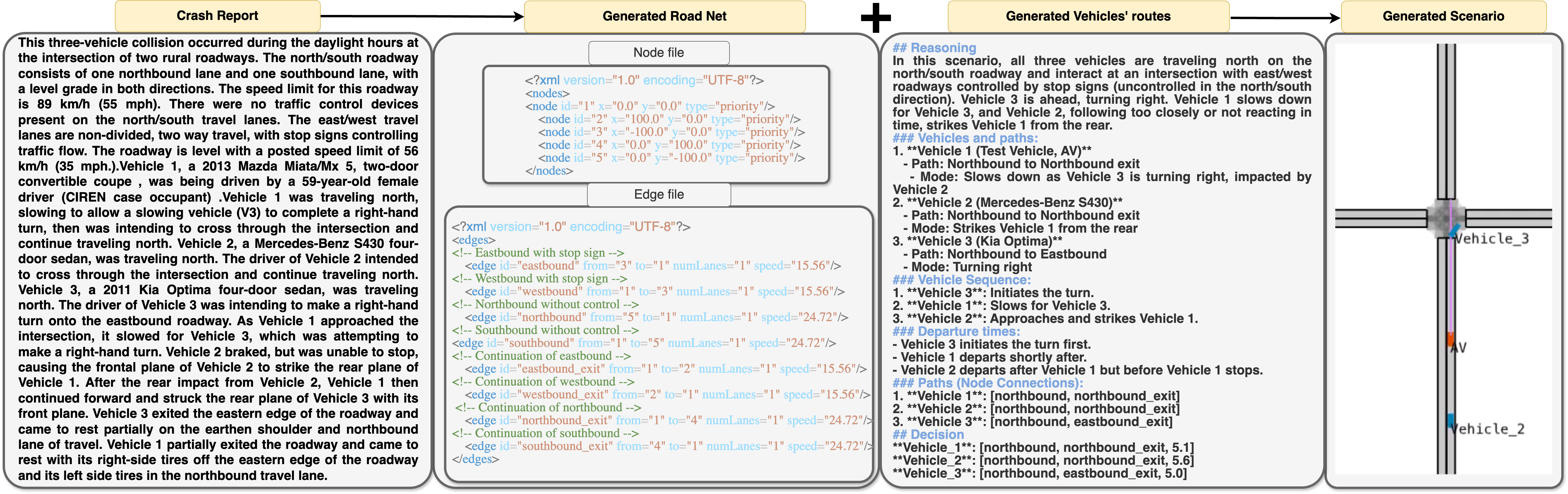}
  \caption{
Based on the crash report, our framework can generate the road network using all the provided information. It can also infer the vehicles' driving routes before the accident, thereby reconstructing the entire scenario.\label{fig:crash_report}}  
\end{figure*}

\begin{figure*}[htbp]  
  \centering  
\includegraphics[width=0.79\textwidth]{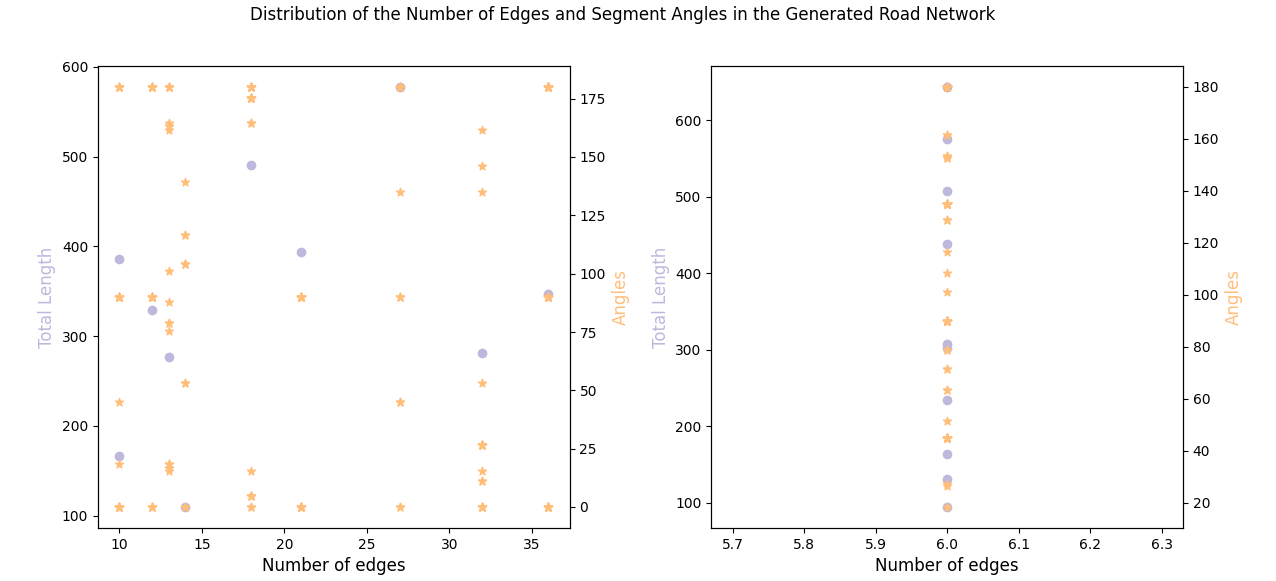}
  \caption{
Left: Diverse road networks generated with varying numbers of edges, segment angles, and total lengths. Right: Similar road networks with the same number of edges due to the removal of the interpreter component.\label{fig:net_scatter}}  
\end{figure*}

\subsection{Effectiveness of OmniTester with crash report}

A direct application of our OmniTester framework is reconstructing scenarios similar to those described in text. This allows for the generation of multiple challenging testing scenarios that can be used for stress testing AV systems. For example, by varying the departure time for each vehicle or changing the driving strategies, a wide range of testing scenarios near risky situations can be generated. 

As an example, we use one case from crash reports \cite{NHTSA2023} to demonstrate how to reconstruct similar dangerous situations for testing a vehicle's reaction. Specifically, instead of using the Interpreter, we directly use the scenario description from crash report for Net Generator as well as the Vehicle Generator. As shown in Fig.\ref{fig:crash_report}, the text is extracted directly from a crash report, describing a crash event that occurred near an intersection. It includes the moving direction for the roads and their speed limits in km/h. The Net Generator not only generates the exact geometry as described in the report but also automatically converts the speed unit in SUMO's default speed unit of m/s without error. For the Vehicle Generator, according to the reasoning process present, it can extract each vehicle's property (such as brand and color) and infer their relative positions based only on crash evolution process: "As Vehicle 1 approached the intersection, it slowed for Vehicle 3, which was attempting to make a right-hand turn. Vehicle 2 braked, but was unable to stop, causing the frontal plane of Vehicle 2 to strike the rear plane of Vehicle 1. After the rear impact from Vehicle 2, Vehicle 1 then continued forward and struck the rear plane of Vehicle 3 with its front plane.". Eventually, the system can reconstruct the routes with corresponding departing time to ensure the relative positions match and similar scenario can be reconstructed.

\subsection{Ablation Study}
\label{sec:ablation}
\begin{table}[!h]
\caption{Ablation study: Change of success rate\label{tab:ablation}}
\centering
\begin{tabular}{|c|c|}
\hline
Metrics & Success rate*  \\
\hline
Ours & 1 \\
\hline
without netconvert  &0  \\
\hline
without interpreter & 1 \\
\hline
without reasoning section &0.4 \\
\hline
\end{tabular}
\end{table}

As shown in Table \ref{tab:ablation}, the success rate of the generation process (with a maximum of 3 attempts per scene) drops from 1 to 0 for the generation process when the netconvert component is removed. Several recurrent errors occur, indicating a fundamental inability to generate the network directly. For example, the LLM's response includes unknown keywords in the net file, such as "limitedTurnSpeed=true," resulting in unsuccessful parsing into SUMO. Additionally, the generated net XML is missing junctions and connections and contains incorrect connections, leading to errors like "An unknown lane."

After removing the interpreter, which is responsible for generating detailed narrative descriptions for each scenario, the LLM fails to produce diverse network structures in one pass. It consistently generates 4 nodes with minor spatial variations and 6 edges defined by the same nodes. As shown in Fig.\ref{fig:net_scatter}, compared to the left figure, the right figure lacks diversity in network structures. This demonstrates that diversity is significantly supported by the multi-stage generation process.

After removing the CoT mechanism, the observed degradation in accuracy can be primarily attributed to three prevalent issues. The first issue arises from the omission of critical attributes within the definition of elements. For instance, the absence of the "shape" attribute within the definition of the "lane" element results in SUMO's inability to accurately load and interpret the lane configurations. The second issue stems from the use of attribute values that fall outside the specified enumeration. For example, the "spreadType" attribute in the "edge" element, which indicates how to calculate the lane geometry from the edge geometry, was incorrectly assigned the value "left," although the permissible values are limited to "right," "center," and "roadCenter." The third issue is the utilization of undeclared attributes within elements. For instance, the generated XML file references an incorrect XML Schema, resulting in the "function" attribute within the "edge" element being undeclared.

\section{Conclusion and Future work}\label{sec:conclusion}
 
We present a scenario generation framework based on LLM and VLM models. This is the first system to generate road structures and vehicles based on user requests with high controllability and flexibility, powered by RAG and self-improvement techniques. We demonstrate that a traffic flow simulator can serve as an efficient tool for LLM to generate complex and diverse testing scenarios with properly designed prompts. A structured definition written in XML, combined with image-based representation and narrative description, effectively represents a scenario. Lastly, self-improvement through effective feedback can enhance scenario generation.

Current endeavors are still in their early stages, and utilizing multimodal large language models and the vast array of knowledge learned within these models to create more complex scenarios remains an open question. One direction for future work is to generate more realistic BV behaviors which could be controlled by deep neural networks (DNNs) or another LLM model. Another direction is to model more elements within the scenario based on detailed descriptions. By pursuing this path, we aim to generate controllable world models that enhance the realism and complexity of simulated environments, making testing more targeted and efficient.

\appendices

\section{Prompt Examples} 
\label{sec:appendix}
\begin{figure}[!h]  
  \centering  
\includegraphics[width=0.48\textwidth]{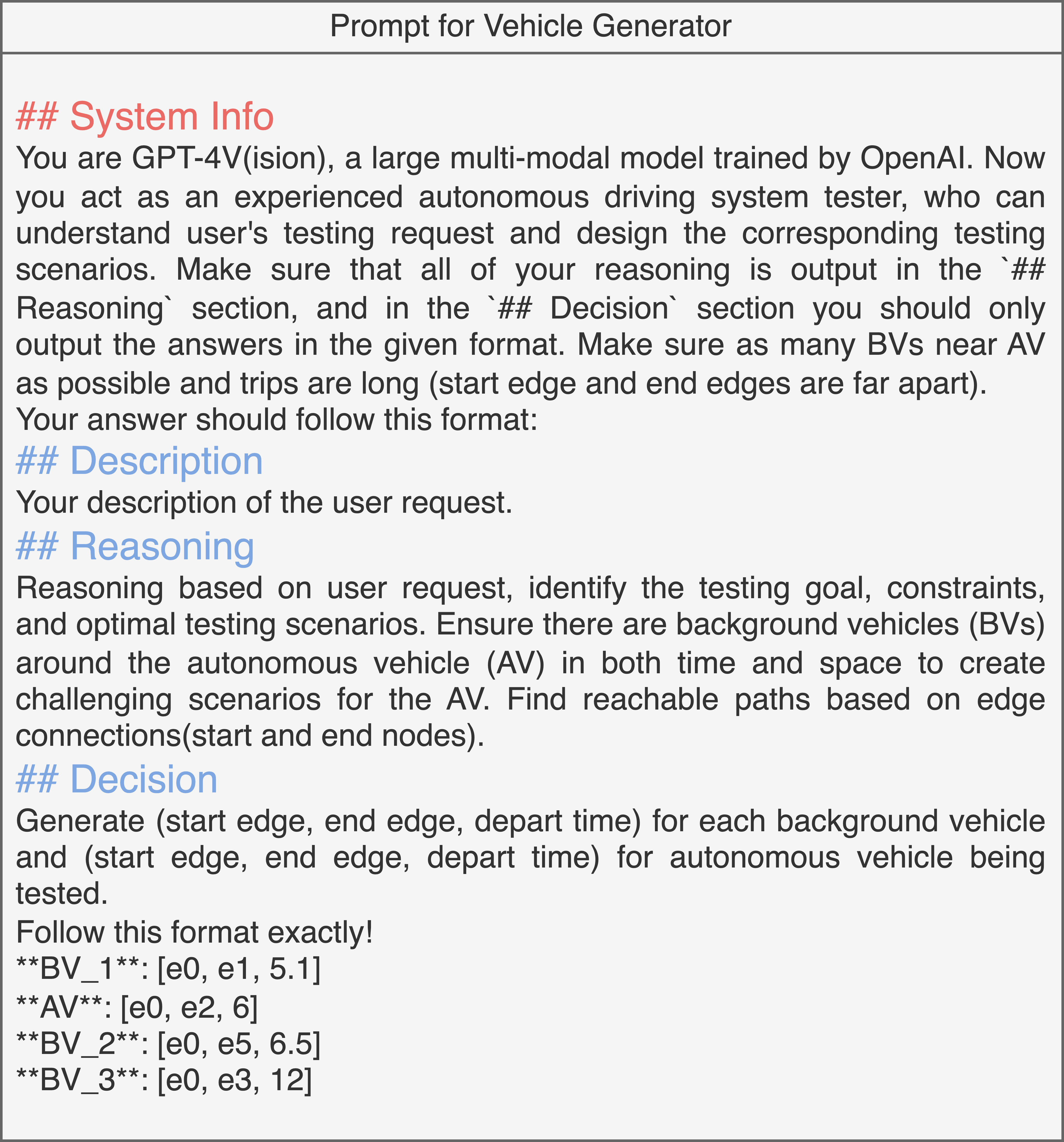}
  \caption{
The system prompts for Net Generator and Vehicle Generator of OmniTester.\label{fig:prompt_comb}}  
\end{figure}

Prompts are crucial in enhancing the reasoning and generation capabilities of LLMs. Fig.\ref{fig:prompt_comb} demonstrate the prompts used in the Net Generator and Vehicle Generator of OmniTester. Each prompt includes several parts: a summary of the generation task, the steps guiding the generation process, and the desired format of the output. For the Net generator, the prompt additionally includes several road constraints to ensure the generated node and edge files are well-defined XML format. For Vehicle Generator, the prompt includes explanations for challenging situations. 
\begin{figure}[!h]  
  \centering  
\includegraphics[width=0.48\textwidth]{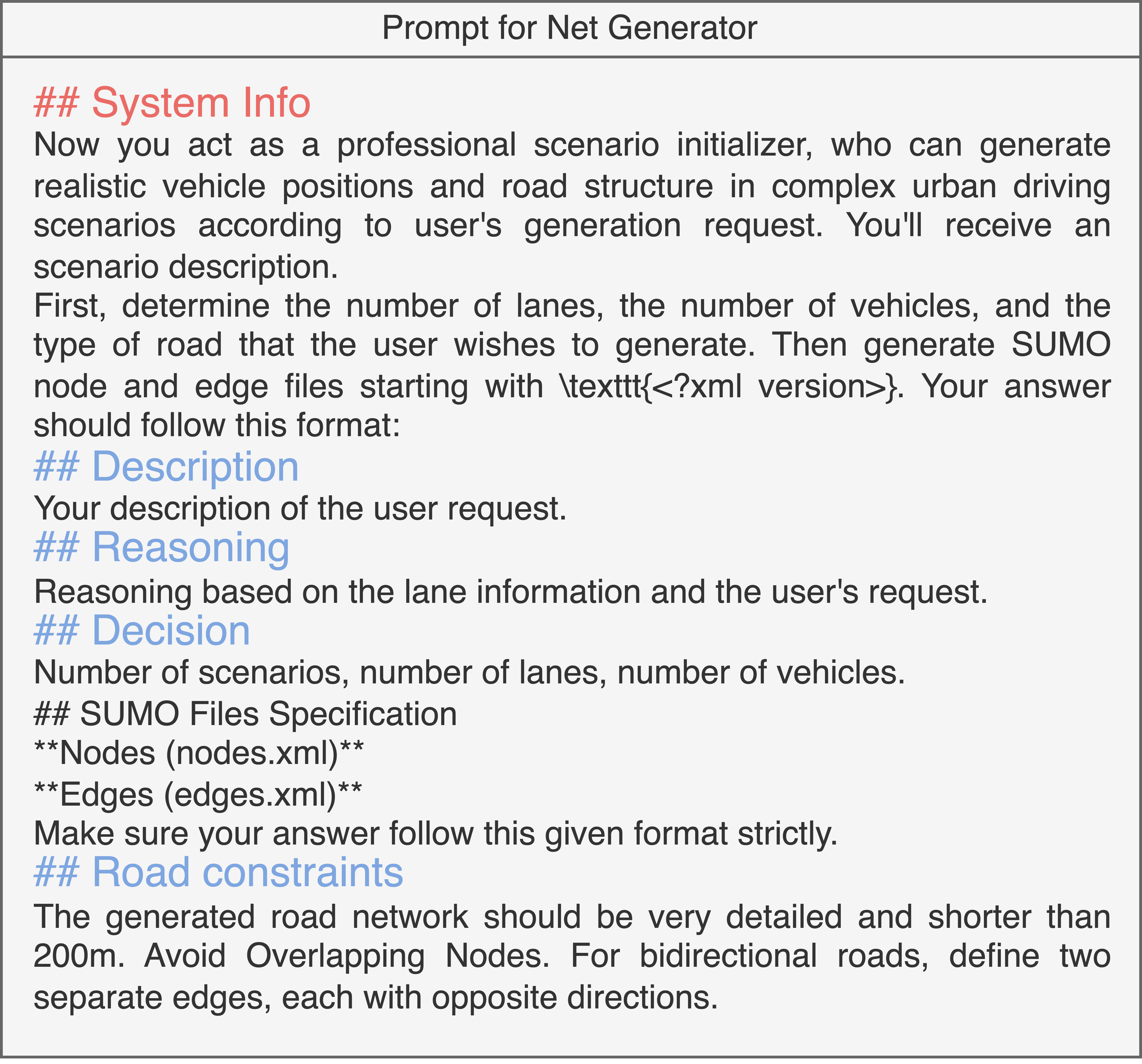}
\end{figure}

{
    \bibliographystyle{IEEEtran}
    \bibliography{main}
}

\end{document}